\documentclass[runningheads]{llncs}

\usepackage{graphicx}
\usepackage{xcolor}
\usepackage{stackengine}
\usepackage{array}
\usepackage{arydshln}
\usepackage{multirow}
\usepackage{pgfplots}
\usepackage{pgfplotstable}
\usepackage{amsmath}
\usepackage{hyperref}
\usepackage{soul}
\usepackage{caption}
\usepackage{subcaption}

\newcolumntype{P}[1]{>{\centering\arraybackslash}p{#1}}

\newenvironment{tableth}{%
		\begin{table}[htbp]
			\centering
	}{
		\end{table}
		}

\begin{document}
\title{Deep learning based registration using spatial gradients and noisy segmentation labels}
\titlerunning{Multi organs registration}
\author{Théo Estienne\inst{1,2} \and Maria Vakalopoulou\inst{1} \and Enzo Battistella\inst{1,2} \and Alexandre Carré\inst{2} \and Théophraste Henry\inst{2} \and Marvin Lerousseau \and Charlotte Robert\inst{2} \and Nikos Paragios\inst{3} \and Eric Deutsch\inst{2}}
\authorrunning{T. Estienne et al.}
\institute{Université Paris-Saclay, CentraleSupélec, Mathématiques et Informatique pour la Complexité et les Systèmes, Inria Saclay, 91190, Gif-sur-Yvette, France. \email{\{theo.estienne\}@centralesupelec.fr} \and Université Paris-Saclay, Institut Gustave Roussy, Inserm, Radiothérapie Moléculaire et Innovation Thérapeutique, 94800, Villejuif, France. \and Therapanacea, Paris, France\\}
\maketitle              % typeset the header of the contribution
\begin{abstract}
Image registration is one of the most challenging problems in medical image analysis. In the recent years, deep learning based approaches became quite popular, providing fast and performing registration strategies. In this short paper, we summarise our work presented on Learn2Reg challenge 2020. The main contributions of our work rely on \textit{(i)} a symmetric formulation, predicting the transformations from source to target and from target to source simultaneously, enforcing the trained representations to be similar and \textit{(ii)} integration of variety of publicly available datasets used both for pretraining and for augmenting segmentation labels. Our method reports a mean dice of $0.64$ for task 3 and $0.85$ for task 4 on the test sets, taking third place on the challenge.  Our code and models are publicly available at \url{https://github.com/TheoEst/abdominal\_registration} and

\url{https://github.com/TheoEst/hippocampus\_registration}. 
\end{abstract}
\section{Introduction}
In the medical field, the problem of deformable image registration has been heavily studied for many years. The problem relies on establishing the best dense voxel-wise transformation ($\Phi$) to wrap one volume (source or moving, $M$) to match another volume (reference or fixed, $F$) in the best way. Traditionally, different types of formulations and approaches had been proposed in the last years~\cite{sotiras2013deformable} to address the problem. However, with the recent advances of deep learning, a lot of learning based methods became very popular currently, providing very efficient and state-of-the art performances~\cite{haskins2020deep}. Even if there is a lot of work in the field of image registration there are still a lot of challenges to be addressed. In order to address these challenges and provide common datasets for the benchmarking of learning based~\cite{voxelmorph,de2019deep} and traditional methods~\cite{heinrich2013mrf,avants2008symmetric}, the Learn2Reg challenge is organised~\cite{adrian_dalca_2020_3715652}. Four tasks were proposed {\color{black} by the organisers} with different organs and modalities. In this work, we focused on two tasks: the CT abdominal (task 3) and the MRI hippocampus registration (task 4).

In this work, we propose a learning based method that learns how to obtain spatial gradients in a similar way to~\cite{stergios2018linear,estienne2020deep}. The main contributions of this work rely on \textit{(i)} enforcing the same network to predict both  $\Phi_{M\rightarrow F}$ and  $\Phi_{F\rightarrow M}$ deformations using the same encoding and implicitly enforcing it to be symmetric and \textit{(ii)} 
{\color{black}integrating noisy labels from different organs during the training, to fully exploit publicly available datasets.} In the following sections, we will briefly summarise these two contributions and present our results that gave to our method the third position in the Learn2Reg challenge 2020 (second for task 3 and third for task 4).

\section{Methodology}
\begin{figure}[t!]
    \centering
    \includegraphics[trim=4.6cm 10.3cm 1.65cm 9.6cm, clip, height=6cm]{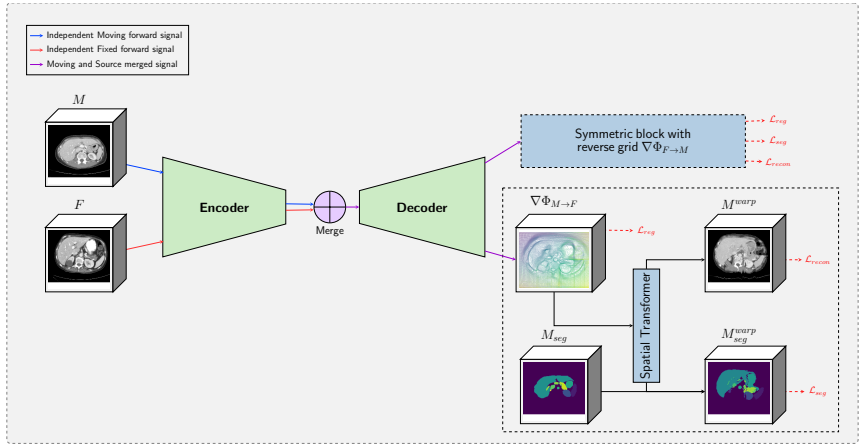}
    \caption{\color{black} Schematic representation of the proposed methodology.}
    \label{fig:network}
\end{figure}

{\color{black} An overview of our proposed framework is presented in the Figure~\ref{fig:network}.} Our method uses as backbone a 3D UNet{\color{black}~\cite{cciccek20163d}} based architecture, which consists of 4 blocks with 64, 128, 256 and 512 channels for the encoder part ($\textbf{E}$).  Each block consists of a normalisation layer, Leaky ReLU activation, $3$D convolutions with a kernel size of $3\times3\times3$ and convolution with kernel size and stride 2 to reduce spatial resolution. Each of the $F,M$ volumes passes independently through the encoder part of the network. Their encoding is then merged using the subtraction operation before passing through the decoder ($\textbf{D}$) part for the prediction of the optimal spatial gradients of the deformation field $\nabla \Phi$. {\color{black}We obtained the deformation field $\Phi$ from  its gradient using integration which we approximated with the cumulative summation operation.} $\Phi$ is then used to obtain the deformed volume together with its segmentation mask using warping  $M^{warp} = \mathcal{W}(M, \Phi_{M\rightarrow F})$. Finally, we apply deep supervision to train our network in a way similar to~\cite{krebs2019learning}.

%\begin{figure}
%    \centering
%    \includegraphics[height=6cm]{Images/method.PNG}
%    \caption{\color{black} Architecture}
%    \label{fig:network}
%\end{figure}

\vspace{-0.2cm}
\paragraph{\textbf{Symmetric training}}
Even if our grid formulation has constraints for the spatial gradients to avoid self-crossings on the vertical and horizontal directions for each of the x,y,z-axis, our formulation is not diffeomorphic. This actually indicates that we can not calculate the inverse transformation of $\Phi_{M\rightarrow F}$. To deal with this problem, we predict both $\Phi_{M\rightarrow F}$ and $\Phi_{F\rightarrow M}$ and we use both for the optimization of our network. Different methods such as~\cite{kim2019unsupervised,guo2020end} explore similar concepts using however different networks for each deformation. Due to our fusion strategy on the encoding part, our approach is able to learn both transformations with less parameters. In particular, our spatial gradients are obtained by: $\nabla\Phi_{M\rightarrow F} = \mathbf{D}( \mathbf{E}(M) - \mathbf{E}(F))$ and $\nabla\Phi_{F\rightarrow M} = \mathbf{D}( \mathbf{E}(F) - \mathbf{E}(M))$.

\vspace{-0.2cm}
\paragraph{\textbf{Pretraining and Noisy Labels}}
Supervision has been proved to boost the performance of the learning based registration methods integrating implicit anatomical knowledge during the training procedure. For this reason, in this study, we investigate ways to use publicly available datasets to boost performance. We exploit available information from publicly available datasets namely KITS 19~\cite{kits_dataset}, Medical Segmentation Decathlon (sub-cohort Liver, Spleen, Pancreas, Hepatic Lesion and Colon)~\cite{task4_dataset}  and TCIA Pancreas\cite{roth2016data,gibson2018automatic}. In particular, we trained a 3D UNet segmentation network on $11$ different organs (spleen, right and left kidney, liver, stomach, pancreas, gallbladder, aorta, inferior vena cava, portal vein and oesophagus). To harmonise the information that we had at disposal for each dataset, we optimised the dice loss only on the organs that were available per dataset. The network was then used to provide labels for the $11$ organs for approximately $600$ abdominal scans. These segmentation masks were further used for the pretraining of our registration network for the task 3. After the training the performance of our segmentation network on the validation set in terms of dice is summarised to: 0.92 (Spl), 0.90 (RKid), 0.91 (LKid), 0.94 (Liv) 0.83 (Sto), 0.74 (Pan), 0.72 (GBla), 0.89 (Aor), 0.76 (InfV), 0.62 (PorV) and 0.61 (Oes). The validation set was composed of 21 patients of Learn2Reg and TCIA Pancreas dataset.

Furthermore, we explored the use of pretraining of registration networks on domain-specific large datasets. In particular, for task 3 the ensemble of the publicly available datasets together with their noisy segmentation masks were used to pretrain our registration network, after a small preprocessing including an affine registration step using  Advanced Normalization Tools (ANTs)\cite{avants2009advanced} and isotropic resampling to $2$mm voxel spacing. {\color{black}Moreover, for task 4, we performed an unsupervised pretraining  using approximately $750$ T1 MRI from OASIS 3 dataset \cite{oasis_dataset} without segmentations.} For both tasks, the pretraining had been performed for 300 epochs. 

\subsection{Training Strategy and Implementation Details}
To train our network, we used a combination of multiple loss functions. The first one was the reconstruction loss optimising a similarity function over the intensity values of the medical volume $\mathcal{L}_{sim}$. For our experiments, we used the mean square error function and normalized cross correlation, depending on the experiment, between the warped image $M^{warp}$ and the fixed image $F$. The second loss integrated anatomical knowledge by optimising the dice coefficient between the warped segmentation and the segmentation of the fixed volume: $\mathcal{L}_{sup} = Dice(M^{warp}_{seg},F_{seg})$. Finally, a regularisation loss was also integrated to enforce smoothness of the displacement field by keeping it close to zero deformation  : $\mathcal{L}_{smo} = || \nabla \Phi_{M\rightarrow F} ||$. These losses composed our final optimization strategy calculated for both $\nabla\Phi_{M\rightarrow F}$ and $\nabla \Phi_{F\rightarrow M}$ 
\begin{equation*}
    \mathcal{L} = (\alpha \mathcal{L}_{sim} + \beta \mathcal{L}_{sup} + \gamma \mathcal{L}_{smo})_{{M\rightarrow F}} + (\alpha \mathcal{L}_{sim} + \beta \mathcal{L}_{sup} + \gamma \mathcal{L}_{smo})_{{F\rightarrow M}}
\end{equation*}

where $\alpha$, $\beta$ and $\gamma$ were weights that were manually defined. 
The network was optimized using Adam optimiser with a learning rate set to $1e^{-4}$.

Regarding the implementation details, for task 3, we used batch size 2 with patch size equal to $144\times144\times144$ due to memory limitations. Our normalisation strategy included the extraction of  three CT windows, which all of them are used as additional channels and min-max normalisation to be in the range $(0,1)$. 
For our experiments we did not use any data augmentation and we set  $\alpha=1$, $\beta=1$ and $\gamma=0.01$. The network was trained on 2 Nvidia Tesla V100 with 16 GB memory, for 300 epochs for $\approx$ 12 hours. 
For task 4, the batch size was set to $6$ with patches of size $64\times64\times64$ while data augmentation was performed by random flip, random rotation and translation. Our normalisation strategy in this case included: $\mathcal{N}(0,1)$ normalisation, clipping values outside of the range $[-5, 5]$ and min-max normalisation to stay to the range $(0,1)$. The weights were set to $\alpha=1$, $\beta=1$ and $\gamma=0.1$ and the network was trained on 2 Nvidia GeForce GTX 1080 GPUs with 12 GB memory for 600 epochs for $\approx$ 20 hours.

{\color{black}The segmentation network, used to produce noisy segmentations, was a 3D UNet trained with batch size $6$,  learning  rate  $1e^{-4}$,  leaky  ReLU  activation  functions,  instance  normalisation layers  and  random  crop  of  patch  of  size  $144\times144\times144$.  During inference, we kept the ground truth segmentations of the organs available, we applied a normalisation with connected components and we  checked  each  segmentations  manually  to  remove  outlier  results.}

\section{Experimental Results}
%The last layer of our network is a convolution with three channels as output corresponding to the deformation along the x, y and z-axis and a sigmoid activation function. In order to help the start of the training, we initialise the last convolution with zeros weights and zeros bias which correspond to the identity displacements. 
%Moreover, we apply deep supervision in a way similar to~\cite{krebs2019learning}. %It is implemented in the form of 4 convolutions layer with input channels 512, 256, 128 and 64 and 3 output channels and 4 upsample layer. We pass the output of each block of the decoder through one convolution, and one upsample layer such that we obtained a grid on the same resolution than the initial images. Then we applied the warped operation, and we calculated the loss to the four intermediate images.

\begin{figure}[t!]
    \centering
     \begin{subfigure}[b]{0.48\textwidth}
         \centering
        \includegraphics[height=2.7cm]{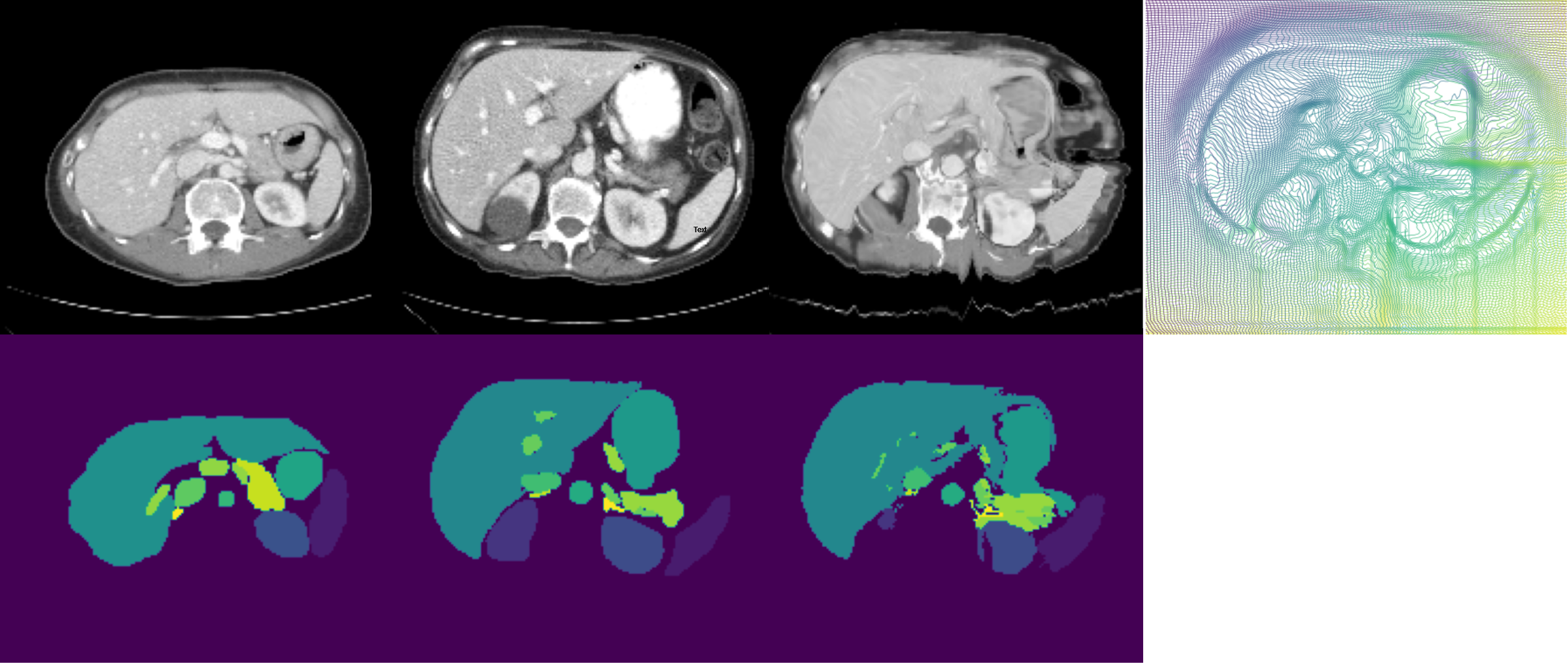}
        \caption{Example for task 3}
         \label{fig:task3}
     \end{subfigure}
     \hfill
     \begin{subfigure}[b]{0.48\textwidth}
         \centering
    \includegraphics[height=2.7cm]{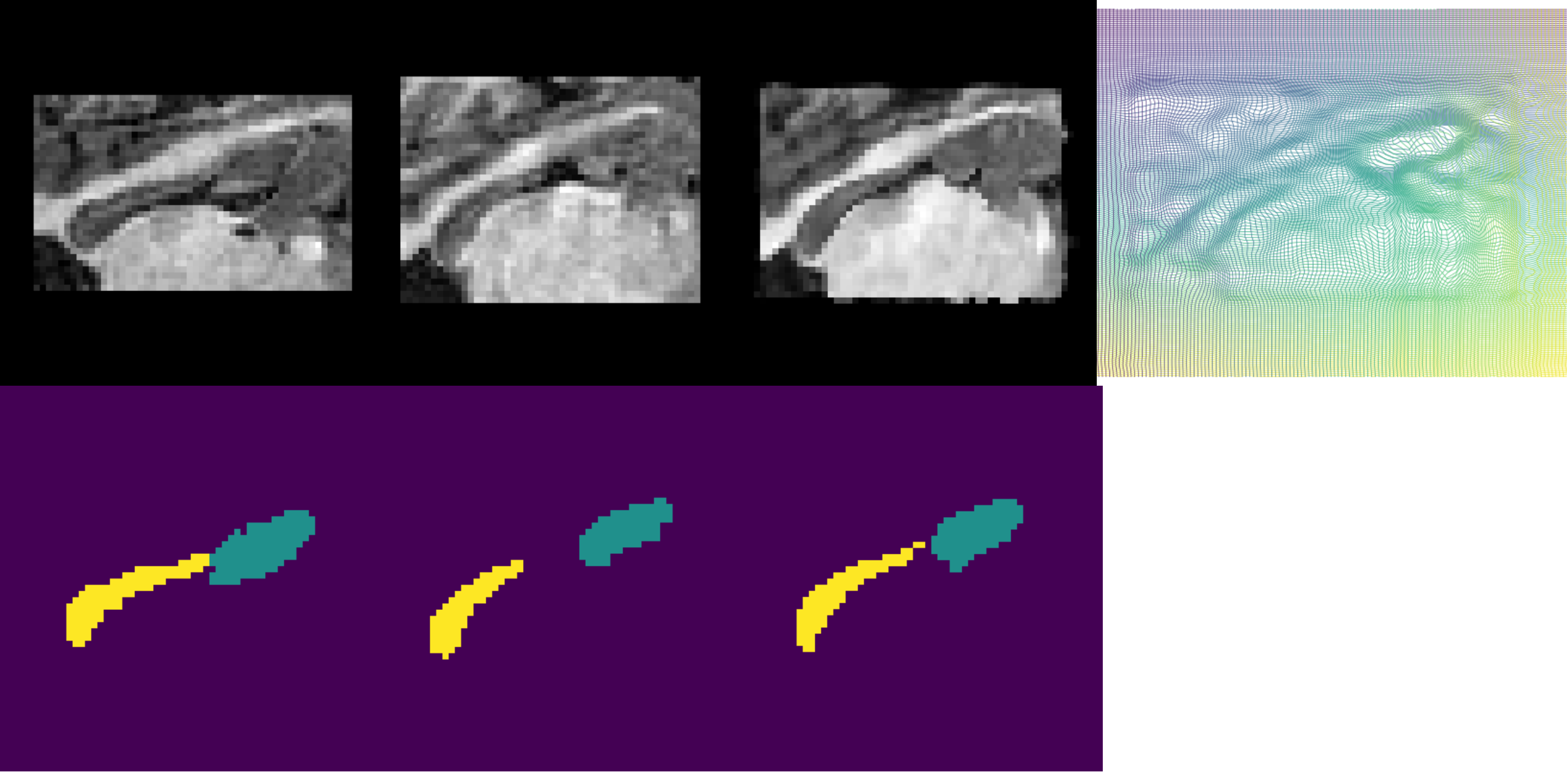}
    \caption{Example for task 4}
    \label{fig:task4}
     \end{subfigure}
        \caption{\color{black}Results obtained on the validation set. From left to right : moving, fixed, deformed images and the deformation grid. For the task 3, we displayed an axial view with the different organs (second row). For the task 4, we displayed a sagittal view with the head and tail masks (second row)}
        \label{fig:results}
\end{figure}

For each task, we performed an ablation study to evaluate the contribution of each component and task 3, we performed a supplementary experiment integrating the noisy labels during the pretraining. The evaluation was performed in terms of Dice score, 30\% of lowest Dice score, Hausdorff distance and standard deviation of the log Jacobian. These metrics evaluated the accuracy and robustness of the method as well as the smoothness of the deformation. Our results are summarised in Table~\ref{tab:task3_val}, {\color{black}while some qualitative results are represented in Figure~\ref{fig:results}. For the inference on the test set, we used our model trained on both training and validation datasets}. Concerning the computational time, our approach needs $6.21$ and  $1.43$ seconds for the inference respectively for task 3 and 4. {\color{black} This is slower than other participants to the challenge, probably due to the size of our deep network which have around 20 millions parameters}.

Concerning task 3, one can observe a significant boost on the performance when the pretraining with the noisy labels was integrated. Due to the challenging nature of this registration problem, the impact of the symmetric training was not so high in any of the metrics. On the other hand, for task 4, the symmetric component with the pretraining boosted the robustness of the method while the pretraining had a lower impact than on task 3. One possible explanation is that for this task, the number of provided volumes in combination with the nature of the problem was enough for training a learning based registration method. 

\begin{tableth}
\scalebox{0.83}{
\begin{tabular}{p{0.1\linewidth}|p{0.50\linewidth}|P{0.05\linewidth}P{0.07\linewidth}P{0.07\linewidth}P{0.06\linewidth}||P{0.05\linewidth} P{0.07\linewidth}P{0.07\linewidth}P{0.06\linewidth}}
& & \multicolumn{4}{c||}{Task 3} & \multicolumn{4}{c}{Task 4}\\
Dataset & & Dice & Dice30 & Hd95 & StdJ & Dice & Dice30 & Hd95  & StdJ \\ \hline
Val & Unregistered & 0.23 & 0.01 &  46.1 & & 0.55 & 0.36 &  3.91 & \\ \hline
Val & Baseline & 0.38 & 0.35 &   45.2 & 1.70 & 0.80 & 0.78 &   2.12 & \textbf{0.067}\\ 
Val & Baseline + sym.  & 0.40 & 0.36 & 45.7 & 1.80 & 0.83 & 0.82 & 1.68 &  0.071 \\ 
Val & Baseline + sym. + pretrain  & 0.52 & 0.50 &  42.3 & \textbf{0.32} & \textbf{0.84} & \textbf{0.83} &  \textbf{1.63}  & 0.093 \\
Val & Baseline + sym. + pretrain  + noisy labels & \textbf{0.62} & \textbf{0.58} &   \textbf{39.3}  & 1.77 & & &\\ \hline
Test & Baseline + sym. + pretrain  + noisy labels & 0.64 & 0.40 & 37.1 & 1.53 & 0.85 & 0.84 & 1.51 & 0.09 

\end{tabular}}
\caption{Evaluation of our method for the Tasks 3 \& 4 of Learn2Reg Challenge {\color{black} on the validation set (val) and on the test set (test)}.}
\label{tab:task3_val}
\end{tableth}

%\begin{tableth}
%\begin{tabular}{p{0.35\linewidth} |P{0.12\linewidth} P{0.12\linewidth} P{0.12\linewidth} P{0.12\linewidth} }  
%& Dice & Dice30 & Hausdorff  & Jacobian \\ \hline
%No Registration & 0.55 & 0.36 &  3.91 & \\ 
%Baseline & 0.796 & 0.777 &   2.12 & \textbf{0.017}\\ 
%Baseline + sym.  & 0.830 & 0.818 & 1.68 &  0.018 \\ 
%Baseline + sym. + pretrain  & \textbf{0.839} & \textbf{0.827} &  \textbf{1.63}  & 0.024\\  \hline
%Test & 0.85 & 0.84 & 1.51 & 0.09 
%\end{tabular}
%\caption{Results for the Task 4}
%\label{tab:task4_val}
%\end{tableth}

\section{Conclusions}
In this work, we summarise our method that took the 3rd place in the  Learn2Reg challenge, participating on the tasks 3 \& 4. Our formulation is based on spatial gradients and explores the impact of symmetry, pretraining and integration of public available datasets. In the future, we aim to further explore symmetry in our method and investigate ways that our formulation could hold diffeomorphic properties. Finally, adversarial training is also something that we want to explore in order to be deal with multimodal registration.
 
\clearpage
\bibliographystyle{plain}
\bibliography{JNAbrv.bib,book}

\end{document}